\documentclass[10pt,twocolumn,letterpaper]{article}

\usepackage[pagenumbers]{cvpr} 

\definecolor{cvprblue}{rgb}{0.21,0.49,0.74}
\usepackage[pagebackref,breaklinks,colorlinks,allcolors=cvprblue]{hyperref}

\usepackage[accsupp]{axessibility} 
\usepackage{amsmath}
\usepackage{bm}
\usepackage{booktabs}
\usepackage{colortbl}
\usepackage{makecell}
\usepackage{subcaption}

\definecolor{Gray}{gray}{0.90}

\newcommand{\sref}[1]{\S\ref{#1}}
\newcommand{\sssection}[1]{\noindent\textbf{#1}}

\def\paperID{22} 
\def\confName{CVPR}
\def\confYear{2026}

\title{ProFashion: Prototype-guided Fashion Video Generation with\\Multiple Reference Images}

\author{Xianghao Kong$^{1}$~, Qiaosong Qi$^2$~, Yuanbin Wang$^2$~, Biaolong Chen$^2$~, Aixi Zhang$^2$~\thanks{Corresponding author.}~, Anyi Rao$^1$~$^*$\\
\small{$^1$~The Hong Kong University of Science and Technology} \ 
\small{$^2$~Taobao \& Tmall Group}
}

\begin{document}
\maketitle

\begin{abstract}
Fashion video generation aims to synthesize temporally consistent videos from reference images of a designated character. Despite significant progress, existing diffusion-based methods only support a single reference image as input, severely limiting their capability to generate view-consistent fashion videos, especially when there are different patterns on the clothes from different perspectives. Moreover, the widely adopted motion module does not sufficiently model human body movement, leading to sub-optimal spatiotemporal consistency. To address these issues, we propose ProFashion, a fashion video generation framework leveraging multiple reference images to achieve improved view consistency and temporal coherency. To effectively leverage features from multiple reference images while maintaining a reasonable computational cost, we devise a Pose-aware Prototype Aggregator, which selects and aggregates global and fine-grained reference features according to pose information to form frame-wise prototypes, which serve as guidance in the denoising process. To further enhance motion consistency, we introduce a Flow-enhanced Prototype Instantiator, which exploits the human keypoint motion flow to guide an extra spatiotemporal attention process in the denoiser. To demonstrate the effectiveness of ProFashion, we extensively evaluate our method on the MRFashion-7K dataset we collected from the Internet. ProFashion also outperforms previous methods on the UBC Fashion dataset.
\end{abstract}    
\section{Introduction}
\label{sec:intro}

Fashion video generation aims to illustrate various nuances of a designated garment by creating coherent and controllable videos from given reference images of a specified character wearing the garment~\cite{dreampose}. It has tremendous application potential in online retail due to its ability to showcase comprehensive details of the garment and the actual look when wearing the clothes. With the recent advancement of diffusion-based video generation methods~\cite{videocomposer, chen2023videocrafter1, make-your-video, dragnuwa, mcdiff}, the fashion video generation task has attracted an increasing amount of attention from both academia and industry~\cite{lfdm, dreampose, wang2023disco, wang2024leo}.

\begin{figure}[t]
    \centering
    \includegraphics[width=\linewidth]{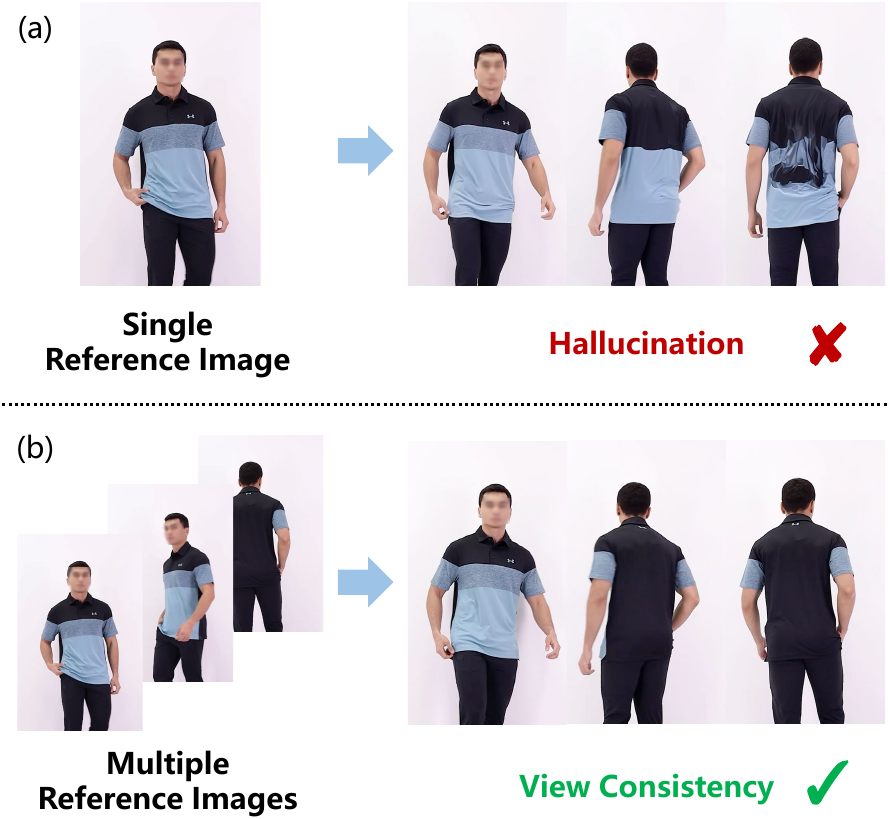}
    \caption{Single reference image fails to provide sufficient information when generating fashion videos for garments with view-dependent patterns and leads to severe hallucination. In contrast, multi-image-conditioned fashion video generation ensures satisfactory view consistency (\sref{sec:intro}).}
    \label{fig:motivation}
\end{figure}

Although significant progress has been made~\cite{dreampose, animateanyone, magicanimate, champ}, previous diffusion-based methods can only accept a single reference image as input, resulting in performance degradation when handling garments with more complex patterns that cannot be depicted by only one reference image. For instance, there are numerous clothes that have different patterns on the front and back sides respectively. As displayed in \cref{fig:motivation}~(a), generating fashion videos showing both sides of such garments with a single reference image as condition will lead to serious hallucination, which is not intended when illustrating clothes to potential customers. Moreover, fashion videos that provide an all-round look of garments typically include large human body movements like turning around. However, current methods mostly adopt a motion module~\cite{guo2024animatediff} that only propagates information on the same spatial position along the temporal dimension, which is insufficient to maintain a satisfactory spatiotemporal consistency when generating fashion videos with substantial body movements.

To address the aforementioned issues, we introduce ProFashion, a prototype-guided fashion video generation framework that can effectively exploit information from multiple reference images to achieve enhanced view consistency and motion stability (\cref{fig:motivation}~(b)).
This paper primarily encompasses the following four technical contributions:
\textbf{First}, to overcome the inherent information limitation of single-image-conditioned fashion video generation, we extend the fashion video generation task to multiple reference images, converting the originally ill-posed task to a more tractable problem by providing reference information from various perspectives.
\textbf{Second}, to provide a reliable and practical solution to multi-image-conditioned fashion video generation, we propose a fashion video generation framework conditioned by multiple reference images and a driving pose sequence. It utilizes a Reference Encoder to extract fine-grained hierarchical features from reference images and a denoiser to incorporate global and fine-grained features from multiple reference images into the denoising process.
\textbf{Third}, to effectively integrate features from multiple reference images into the denoising process without introducing a significant computational burden, we present a Pose-aware Prototype Aggregator, which selects and aggregates global and fine-grained reference features according to pose information to form prototypes for each frame. The aggregated prototypes share the same shape with a single reference feature and thus are capable of guiding the denoising process with the same computational cost as a single reference.
\textbf{Fourth}, to ensure smoothness of character motion and detail consistency across frames, we devise a Flow-enhanced Prototype Instantiator, which incorporates additional spatiotemporal attention layers into the denoiser and leverages the human keypoint motion flow to supervise the spatiotemporal warping process, extending temporal information propagation to other relevant spatial locations.

To validate the effectiveness of ProFashion on multi-image-conditioned fashion video generation, we construct MRFashion-7K, an Internet-collected fashion video dataset containing 7,335 fashion videos with diverse garment details from different perspectives of characters. On this dataset, ProFashion significantly outperforms single-reference baselines in both subjective and objective evaluations. ProFashion also surpasses other state-of-the-art methods on the UBC Fashion~\cite{dwnet_ubc} dataset.

\section{Related Work}
\label{sec:related}

\subsection{Diffusion-based Visual Content Generation}

In recent years, the emergence of diffusion models~\cite{ddpm, ddim} has boosted the advancement of visual content generation due to their higher training stability and better generation diversity. Latent Diffusion Model~\cite{ldm} proposes to perform the diffusion process in a low-dimensional latent space~\cite{vqvae}, striking a balance between generation quality and computational complexity. IP-Adapter~\cite{ipadapter} designs a lightweight structure to adapt text-to-image models to image conditions. ControlNet~\cite{controlnet} introduces an effective way to inject pixel-wise control signals like poses and depths into the denoising process. To leverage the scaling capability~\cite{kaplan2020scalinglawsneurallanguage} of the transformer~\cite{attention2017} architecture, DiT~\cite{dit} substitutes the denoising U-Net~\cite{unet} with a transformer structure, achieving promising generation quality and scaling-up potential. Thanks to these fundamental works, diffusion-based image generation methods~\cite{imagen, ramesh2022hierarchicaltextconditionalimagegeneration, glide, balaji2023ediffitexttoimagediffusionmodels, composer, t2i-adapter} have flourished and achieved satisfactory performance.

Along with the developments in the image domain, researchers have also been trying to lift methods for images up to videos~\cite{vdm}. Compared to 2D images, videos introduce an additional temporal dimension, further challenging the model with complicated inter-frame relationship comprehension and a huge amount of computation~\cite{imagenvideo, singer2022makeavideo}. Most recent works~\cite{imagenvideo, singer2022makeavideo, zhang2024show1, alignyourlatents, svd, t2vzero, zhang2024controlvideo, guo2024animatediff, tune-a-video, hong2023cogvideo, latte} address the above challenges by inserting extra temporal convolution and attention layers to model the temporal relationship while decoupling the expensive and complicated 3D dependencies. Besides text-to-video generation, there have also been methods~\cite{videocomposer, chen2023videocrafter1, make-your-video, dragnuwa, mcdiff} using images as generation conditions. However, these methods can only handle a single reference image, falling short in generating view-consistent videos leveraging reference images from multiple perspectives.

\subsection{Human Video Generation}

\begin{figure*}[ht]
    \centering
    \includegraphics[width=0.93\linewidth]{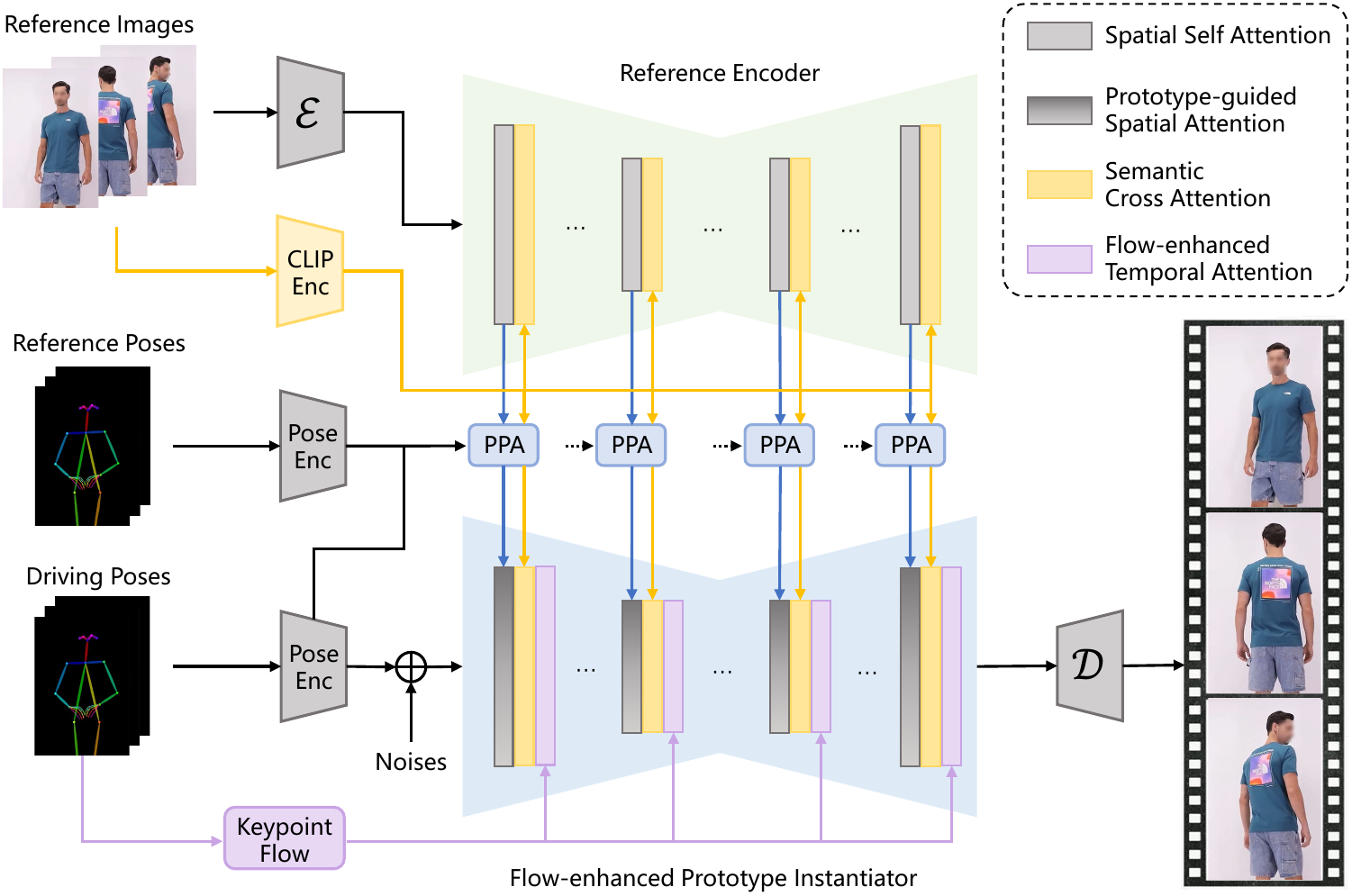}
    \caption{Overall framework of ProFashion (\sref{sec:framework}). It first converts the inputs into latent spaces with different encoders. Then, a Reference Encoder (\sref{sec:ref_enc}) is used for extracting multi-scale representation of reference images. Next, PPA (\sref{sec:ppa}) is adopted to aggregate the multi-scale representation and global features into fine-grained and global prototypes according to pose similarity. Finally, it utilizes FPI (\sref{sec:fpi}) to conduct a prototype-guided iterative denoising process enhanced by keypoint motion flow.}
    \label{fig:framework}
\end{figure*}

Human video generation aims to achieve consistent and controllable human video synthesis based on given reference images~\cite{siarohin2019animating, tpsmm, bdmm, siarohin2021motion, siarohin2019first} or videos~\cite{everybodydancenow}. Due to the promising results and flexible controllability, researchers have been adopting diffusion-based methods to the field of human video generation~\cite{lfdm, dreampose, wang2023disco, wang2024leo}. Animate Anyone~\cite{animateanyone} introduces a method leveraging a ReferenceNet structure to inject the reference image into the denoising U-Net with a spatial attention mechanism. It also adopts a lightweight Pose Guider to control the motion of generated characters. MagicAnimate~\cite{magicanimate} utilizes an appearance encoder network to integrate identity information and a ControlNet~\cite{controlnet} to achieve pose control. It also proposes a sliding window mechanism to achieve long video generation with high spatial consistency. Champ~\cite{champ} utilizes SMPL~\cite{smpl} to achieve more accurate body shape and pose control. It adopts a multi-layer motion fusion module to integrate depth images, normal maps, segmentation maps, as well as skeletons, into the denoising U-Net. Although significant progress has been made, existing methods still struggle to generate view-consistent human videos with diverse clothing and large character movements conditioned by multiple reference images from diverse perspectives.

\section{Methodology}
\label{sec:method}

\subsection{Task Formulation}
\label{sec:formulation}

Given $N_r$ reference images $\bm{I}_r^{1:N_r}$, the corresponding rendered character poses $\bm{p}_r^{1:N_r}$, and the driving pose sequence $\bm{p}^{1:N_f}$ containing $N_f$ rendered poses, the multi-image-conditioned fashion video generation task is to synthesize a coherent video $\bm{I}^{1:N_f}$ in which the character's appearance aligns with $\bm{I}_r^{1:N_r}$ and its motion matches $\bm{p}^{1:N_f}$.

\subsection{Overall Framework}
\label{sec:framework}

To leverage the recent advancements in diffusion-based video generation methods, we build the proposed ProFashion upon a latent diffusion model~\cite{ldm}. The overall structure of ProFashion is illustrated in \cref{fig:framework}. It contains three main components: a Reference Encoder (\sref{sec:ref_enc}), a Pose-aware Prototype Aggregator (PPA, \sref{sec:ppa}), and a Flow-enhanced Prototype Instantiator (FPI, \sref{sec:fpi}). 

The inputs are encoded into latent spaces at the beginning of the generation process. We encode the reference images $\bm{I}_r^{1:N_r}$ using a VAE encoder~\cite{vqvae} $\mathcal{E}$ to obtain the fine-grained latent representations $\bm{z}_r^{1:N_r}$. Global representations $\bm{x}_r^{1:N_r}$ of the reference images $\bm{I}_r^{1:N_r}$ are extracted by a CLIP image encoder~\cite{clip} $\mathcal{E}_{clip}$.
The reference poses $\bm{p}_r^{1:N_r}$ and the driving pose sequence $\bm{p}^{1:N_f}$ are encoded by a lightweight pose encoder $\mathcal{E}_{pose}$ which contains several convolutional layers and shares a similar structure with the condition encoder in ControlNet~\cite{controlnet} to get the encoded pose features $\bm{x}_{rp}^{1:N_r}$ and $\bm{x}_p^{1:N_f}$ respectively.

The Reference Encoder takes the encoded reference images $\bm{z}_r^{1:N_r}$ and $\bm{x}_r^{1:N_r}$ as inputs, extracting a multi-scale fine-grained representation of reference images $\bm{z}_{r,1:N_l}^{1:N_r}$, where $N_l$ stands for the number of internal blocks (\cref{eq:ref_enc}).
\begin{equation}
    \bm{z}_{r,1:N_l}^{1:N_r}=\mathcal{F}_{ref}(\bm{z}_r^{1:N_r}, \bm{x}_r^{1:N_r})
    \label{eq:ref_enc}
\end{equation}

PPA operates at each block of the Reference Encoder respectively, aggregating multi-scale reference representation at the $j$-th block $\bm{z}_{r,j}^{1:N_r}$ and the global representation $\bm{x}_r^{1:N_r}$ into fine-grained and global prototypes. For the $i$-th frame, the aggregation is performed under the guidance of the driving pose $\bm{x}_p^i$ and the reference poses $\bm{x}_{rp}^{1:N_r}$. \cref{eq:ppa_fine_grained,eq:ppa_global} describe these processes:
\begin{equation}
    \bm{z}_{R,j}^i=\mathcal{F}_{PPA-F}(\bm{z}_{r,j}^{1:N_r},\bm{x}_p^i,\bm{x}_{rp}^{1:N_r}),
    \label{eq:ppa_fine_grained}
\end{equation}
\begin{equation}
    \bm{x}_R^i=\mathcal{F}_{PPA-G}(\bm{x}_{r}^{1:N_r},\bm{x}_p^i,\bm{x}_{rp}^{1:N_r}),
    \label{eq:ppa_global}
\end{equation}
where $\mathcal{F}_{PPA-F}$ denotes fine-grained PPA and $\mathcal{F}_{PPA-G}$ denotes global PPA.

FPI conducts an iterative denoising process by predicting noise at each timestep. The driving pose features $\bm{x}_p^{1:N_f}$ are added to the noise latent $\bm{z}^{1:N_f}$ to form the input latent $\bm{z}_0^{1:N_f}$.
During noise prediction, FPI exploits information from fine-grained and local prototypes (\cref{eq:fpi}).
\begin{equation}
    \bm{z}_{pred}^{1:N_f}=\mathcal{F}_{FPI}(\bm{z}_0^{1:N_f},\bm{z}_{R,1:N_l}^{1:N_f},\bm{x}_R^{1:N_f})
    \label{eq:fpi}
\end{equation}

Finally, the denoised latents are converted back to pixel space by a VAE decoder~\cite{vqvae} $\mathcal{D}$ to form a consistent fashion video. We present the training strategy in \sref{sec:training}.

\subsection{Reference Encoder}
\label{sec:ref_enc}

The Reference Encoder is a U-Net-based~\cite{unet} structure for extracting multi-scale fine-grained features of reference images. After each convolution block~\cite{resnet}, it also includes an attention block which consists of a spatial self-attention layer and a semantic cross-attention layer to further enrich the semantic information in the latent representations.

The spatial self-attention layer conducts self-attention on the spatial dimension of reference latents. In the $j$-th attention block, we perform self-attention on the input latent of the $k$-th reference image $\bm{z}_{r,j-1}^k$ to obtain $\bm{z}_{rs,j}^k$.

To take advantage of the visual representation capability of CLIP~\cite{clip}, the semantic cross-attention layer is adopted to inject extra global information into reference latents. After the $j$-th spatial self-attention layer, we conduct cross-attention~\cite{attention2017} between the output latent of the $k$-th reference image $\bm{z}_{rs,k}^i$ and the CLIP visual feature $\bm{x}_{r}^{k}$ to get $\bm{z}_{r,j}^k$,
where $\bm{z}_{rs,j}^k$ is the attention query and $\bm{x}_{r}^{k}$ serves as the attention key and value.

\subsection{Pose-aware Prototype Aggregator (PPA)}
\label{sec:ppa}

\begin{figure}[ht]
    \centering
    \includegraphics[width=\linewidth]{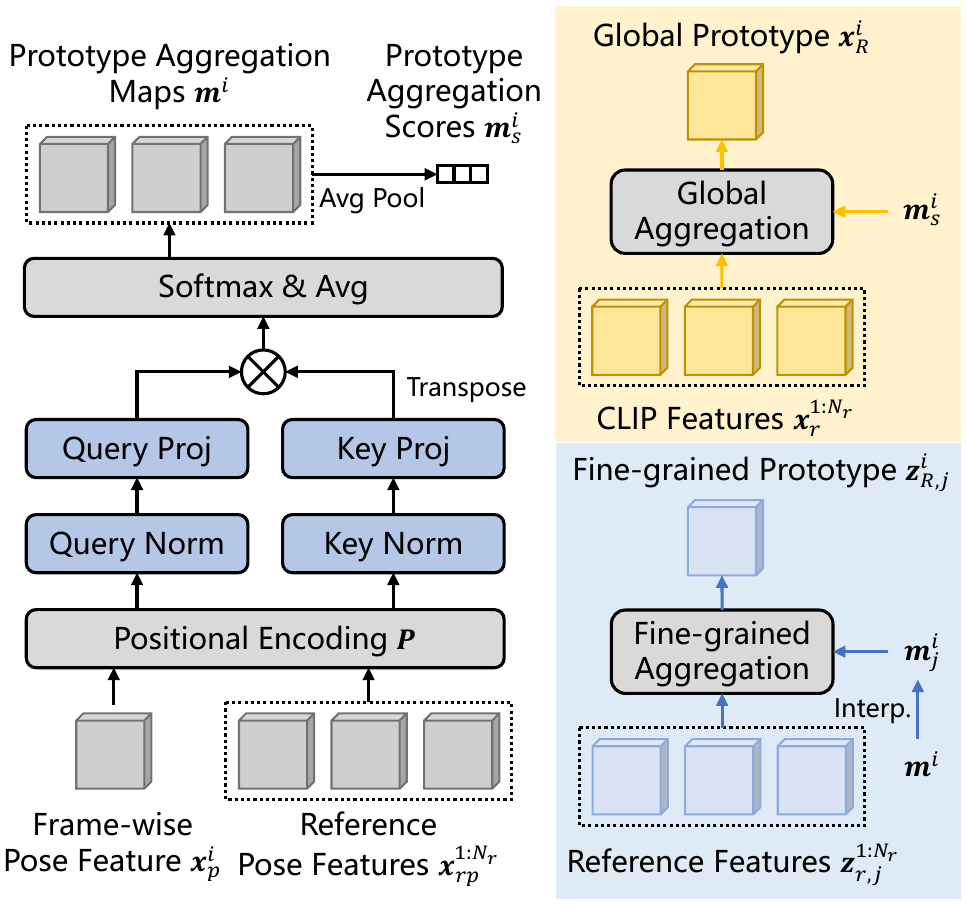}
    \caption{Details of PPA (\sref{sec:ppa}). It first uses a pose-aware selector to calculate the prototype aggregation map and then conducts fine-grained and global prototype aggregation accordingly.}
    \label{fig:ppa}
    \vspace{-0.5em}
\end{figure}

For each frame, PPA aggregates fine-grained features from the Reference Encoder and global features from the CLIP~\cite{clip} visual encoder into prototypes respectively at each attention block, which are subsequently used for guiding the denoising process. The detailed structure of PPA is illustrated in \cref{fig:ppa}.

Intuitively, the reference image whose character pose has a large similarity with the driving pose possesses more information concerning the target view and thus is supposed to account for a more significant proportion in the aggregated prototypes. Inspired by this principle, we design a pose-aware selector to obtain prototype aggregation maps $\bm{m}^i$ according to the similarity of the pose feature of the $i$-th frame $\bm{x}_p^i$ and the reference pose features $\bm{x}_{rp}^{1:N_r}$. The process is explained in the left part of \cref{fig:ppa}. To be specific, we first add sinusoidal positional encoding $\bm{P}$ to the pose features and perform Group Normalization~\cite{groupnorm} and linear projection as \cref{eq:pose_query_proj,eq:pose_key_proj}.
\begin{equation}
    \bm{q}_p^i=\text{\texttt{linear}}(\text{\texttt{group\_norm}}(\bm{x}_p^i+\bm{P}))
    \label{eq:pose_query_proj}
\end{equation}
\begin{equation}
    \bm{k}_p^i=\text{\texttt{linear}}(\text{\texttt{group\_norm}}(\bm{x}_{rp}^{1:N_r}+\bm{P}^{1:N_r}))
    \label{eq:pose_key_proj}
\end{equation}
Then, we perform matrix multiplication, softmax operation, and average pooling between $\bm{q}_p^i$ and $\bm{k}_p^i$ as \cref{eq:pose_multiplication}:
\begin{equation}
    \bm{m}^i=\text{\texttt{avgpool}}(\text{\texttt{softmax}}(\frac{\bm{q}_p^i{\bm{k}_p^i}^T}{\sqrt{d}})),
    \label{eq:pose_multiplication}
\end{equation}
where $d$ is the hidden dimension of features and the average pooling is done on the spatial dimensions of $\bm{q}_p^i$ so that $\bm{m}^i$ can easily conduct Hadamard product with $\bm{z}_{r,j}^{1:N_r}$ to get frame-wise fine-grained prototypes.

For fine-grained aggregation at the $j$-th block, we first perform bilinear interpolation on the original prototype aggregation map $\bm{m}^i$ to get $\bm{m}_j^i$ which shares the same spatial size with $\bm{z}_{r,j}^{1:N_r}$. Then, we conduct Hadamard product between them to obtain the fine-grained prototype as \cref{eq:fine_grained_proto}:
\begin{equation}
    \bm{z}_{R,j}^i=\text{\texttt{sum}}(\bm{z}_{r,j}^{1:N_r}\odot\bm{m}_j^i),
    \label{eq:fine_grained_proto}
\end{equation}
where the sum operation is done on the $1:N_r$ dimension.

As for global aggregation, we perform average pooling on the spatial dimensions of $\bm{m}^i$ to get prototype aggregation scores $\bm{m}_s^i$.
Then, we conduct a weighted sum of global features $\bm{x}_{r}^{1:N_r}$ to obtain the global prototype as \cref{eq:global_proto}:
\begin{equation}
    \bm{x}_R^i=\text{\texttt{sum}}(\bm{x}_{r}^{1:N_r}\odot\bm{m}_s^i),
    \label{eq:global_proto}
\end{equation}
where the sum operation is done on the $1:N_r$ dimension.

The aggregated prototypes contain critical information from all the reference images while sharing the same shape as the features of a single reference image, ensuring effective guiding of the denoising process without introducing subsequent computational burdens.

\subsection{Flow-enhanced Prototype Instantiator (FPI)}
\label{sec:fpi}

FPI instantiates the aggregated prototype through a U-Net~\cite{unet} based denoiser structure according to the driving pose sequence. After each convolution block~\cite{resnet}, it includes an attention block with a Prototype-guided Spatial Attention layer, a semantic cross-attention layer, and a Flow-enhanced Temporal Attention layer.

The Prototype-guided Spatial Attention layer performs cross-attention~\cite{attention2017} on the spatial dimension of latents and fine-grained prototypes. In the $j$-th attention block, we conduct cross-attention between the input latent $\bm{z}_{j-1}^i$ and the spatial concatenation of the input latent $\bm{z}_{j-1}^i$ and the fine-grained prototype $\bm{z}_{R,j}^i$ on the $i$-th frame as \cref{eq:proto_sa}:
\begin{equation}
    \bm{z}_{s,j}^i=\text{\texttt{cross\_attn}}(\bm{z}_{j-1}^i,\bm{z}_{j-1}^i\oplus\bm{z}_{R,j}^i),
    \label{eq:proto_sa}
\end{equation}
where $\bm{z}_{j-1}^i$ is the attention query and the concatenation result acts as the attention key and value.

The semantic cross-attention layer shares a similar structure to that in the Reference Encoder, only substituting the attention key and value for global prototype $\bm{x}_R^i$ of frame $i$.

\begin{figure}[t]
    \centering
    \includegraphics[width=0.93\linewidth]{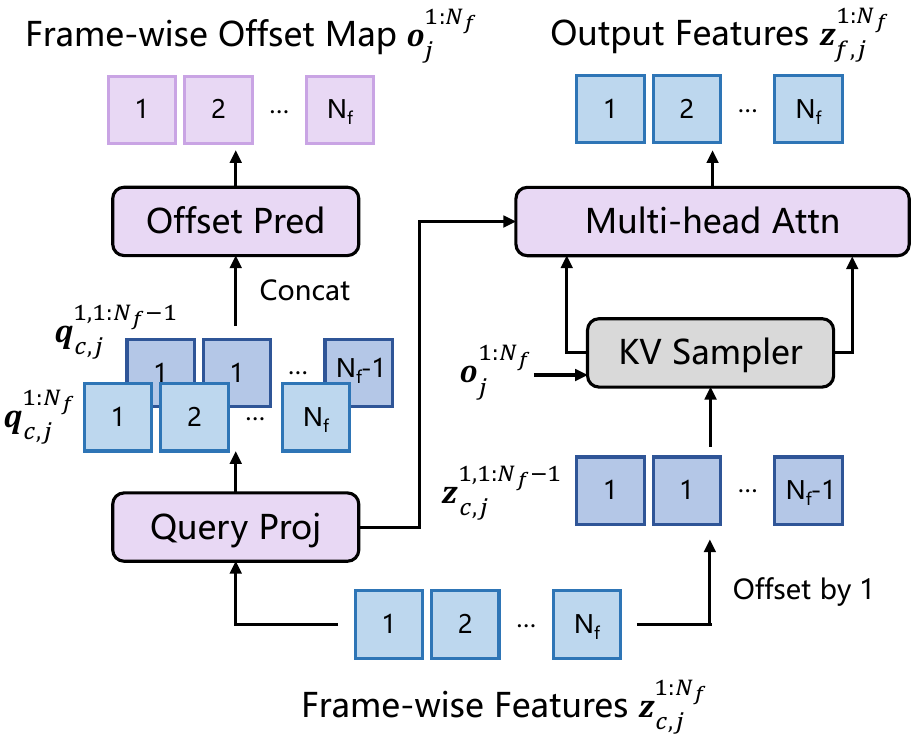}
    \caption{Details of spatiotemporal attention in FTA (\sref{sec:fpi}). It conducts multi-head attention with original frame-wise features as queries and resampled features with 1 frame's offset as keys and values. The resampling process is guided by the query-conditioned offset prediction supervised by human keypoint motion flow.}
    \label{fig:fta}
    \vspace{-1em}
\end{figure}

The Flow-enhanced Temporal Attention (FTA) layer further enhances motion smoothness by introducing an additional spatiotemporal attention process before the prevalently adopted temporal attention layers~\cite{guo2024animatediff}. For fashion videos, the same part of the body is supposed to be consistent across frames. Accordingly, the spatiotemporal attention process is designed to propagate features of the same body part between adjacent frames under the guidance of human keypoint motion flows. The details of this process are depicted in \cref{fig:fta}. It first projects the latents after the semantic cross-attention in the $j$-th layer $\bm{z}_{c,j}^{1:N_f}$ using a linear projection layer to get $\bm{q}_{c,j}^{1:N_f}$.
Then, it concatenates the query of each frame with that of the previous frame along the channel dimension as \cref{eq:flow_query_cat}.
\begin{equation}
    \bm{q}_{cat,j}^{1:N_f}=\bm{q}_{c,j}^{1:N_f}\oplus\bm{q}_{c,j}^{1,1:N_f-1}
    \label{eq:flow_query_cat}
\end{equation}
The concatenated queries are used to predict frame-wise offset maps $\bm{o}_j^{1:N_f}$ with an offset prediction head $\mathcal{F}_{offset}$ which consists of several convolutional layers. The predicted dense offset map $\bm{o}_j^{1:N_f}$ is supervised by the keypoint flow map $\bm{\delta}^{1:N_f}$ extracted from the driving pose sequence $\bm{p}^{1:N_f}$ using Farneback method~\cite{farneback}, which is essentially sparse.
The key and value of the attention process is first obtained by a bilinear sampler $f_{bilinear}$ according to the predicted offset map $\bm{o}_j^{1:N_f}$ from the input latents with offset of 1 frame $\bm{z}_{c,j}^{1,1:N_f-1}$ as \cref{eq:flow_kv_sample}.
\begin{equation}
    \bm{u}_{c,j}^{1:N_f}=f_{bilinear}(\bm{z}_{c,j}^{1,1:N_f-1},\bm{o}_j^{1:N_f})
    \label{eq:flow_kv_sample}
\end{equation}
Then, we apply a multi-head attention~\cite{attention2017} process with $\bm{q}_{c,j}^{1:N_f}$ as query and $\bm{u}_{c,j}^{1:N_f}$ as key and value to get the attention output denoted as $\bm{z}_{f,j}^{1:N_f}$. After the proposed spatiotemporal attention process, we conduct the widely used temporal attention~\cite{guo2024animatediff} along the temporal dimension of $\bm{z}_{f,j}^{1:N_f}$ to get $\bm{z}_j^{1:N_f}$, the final latent output of the $j$-th attention block.

\begin{table*}
  \centering
  \caption{Quantitative results and human evaluation on MRFashion-7K (\sref{sec:mrfashion}). $N_r=3$ unless otherwise specified. Results in \textbf{bold} are the best. \dag~Open-source implementation.}
  \label{tab:mrfashion}
  \resizebox{1\linewidth}{!}{
  \begin{tabular}{l|c|cccc|cccc}
    \toprule
    Settings & \#Frame & \texttt{SSIM}$\uparrow$ & \texttt{PSNR}$\uparrow$ & \texttt{LPIPS}$\downarrow$ & \texttt{FVD}$\downarrow$ & \makecell{Character\\Authenticity} $\uparrow$ & \makecell{Clothing\\Detail} $\uparrow$ & \makecell{Motion\\Fluency} $\uparrow$ & \makecell{Overall\\Quality} $\uparrow$\\
    \midrule
    Animate Anyone\textsuperscript{\dag} ($N_r=1$) & 16 & 0.829 & 20.84 & 0.127 & \cellcolor{Gray}268.50 & 3.44 & 2.50 & 1.81 & \cellcolor{Gray}2.58 \\
    Champ ($N_r=1$) & 16 & 0.831 & 20.88 & 0.126 & \cellcolor{Gray}254.72 & 3.46 & 2.51 & 1.85 & \cellcolor{Gray}2.61 \\
    \midrule
    Baseline ($N_r=1$) & 16 & 0.829 & 21.11 & 0.132 & \cellcolor{Gray}243.98 & 3.45 & 2.53 & 1.86 & \cellcolor{Gray}2.61 \\
    Baseline + Avg Ref & 16 & 0.838 & 21.36 & 0.125 & \cellcolor{Gray}205.45 & 3.94 & 3.13 & 2.88 & \cellcolor{Gray}3.31 \\
    Baseline + Cat Ref & 12 & 0.841 & 22.08 & 0.122 & \cellcolor{Gray}201.89 & 3.95 & 3.18 & 2.82 & \cellcolor{Gray}3.32 \\
    Baseline + PPA & 16 & 0.867 & 23.44 & 0.094 & \cellcolor{Gray}196.95 & 4.31 & 3.69 & 3.12 & \cellcolor{Gray}3.71 \\
    Baseline + PPA + FPI (Ours) & 16 & \textbf{0.885} & \textbf{23.57} & \textbf{0.086} & \cellcolor{Gray}\textbf{126.92} & \textbf{4.56} & \textbf{4.31} & \textbf{3.87} & \cellcolor{Gray}\textbf{4.25} \\
    \bottomrule
  \end{tabular}
  }
\end{table*}

\subsection{Training Strategy}
\label{sec:training}

The training objective of ProFashion $\mathcal{L}$ consists of 2 loss functions. One is the denoising supervision $\mathcal{L}_d$ with the target from v-prediction~\cite{vpred}. The other is the MSE supervision for the offset prediction $\mathcal{L}_o$, in which only non-zero values in $\bm{\delta}^{1:N_f}$ serve as supervision. A hyperparameter $\lambda$ is used for balancing the loss terms as \cref{eq:loss}.
\begin{equation}
    \mathcal{L}=\mathcal{L}_d+\lambda\mathcal{L}_o
    \label{eq:loss}
\end{equation}

We train the proposed ProFashion in 2 stages. In the first stage, we train ProFashion on a single target frame with multiple reference images and exclude all FTA layers. All parameters except those of $\mathcal{E}$, $\mathcal{E}_{clip}$, and $\mathcal{D}$ are updated. In the second stage, the full model is trained on video clips and sequences of repeated still images to enable smooth and consistent motion generation while maintaining the generation quality of individual frames. Only the parameters of FTA layers are updated.

\section{Experiments}
\label{sec:exp}

\subsection{Datasets and Evaluation Metrics}

\sssection{Datasets.} To demonstrate the performance of ProFashion on fashion videos with view-dependent patterns and large motions, we collect MRFashion-7K from the Internet, which contains 7,335 fashion videos with diverse clothing details from different perspectives and significant body movements like turning. There are 6,601 videos for training and 734 videos for testing. Each video is around 10 seconds long. When reporting quantitative results, we select a subset of 16 videos from the test split for evaluation.

For better comparison with other methods, we also evaluate ProFashion on the UBC Fashion~\cite{dwnet_ubc} dataset, which includes 500 fashion videos of about 10 to 15 seconds for training and 100 for testing. We do not use additional data for training.

\sssection{Evaluation Metrics.} We assess the generation quality of ProFashion on both image and video level. For image level evaluation, we use SSIM~\cite{ssim}, PSNR~\cite{psnr}, and LPIPS~\cite{lpips} as quantitative metrics. For video level assessment, we select FVD~\cite{fvd} as the metric.

\subsection{Implementation Details}

\sssection{Reference Image Selection.} During training, the $N_r=3$ reference images of each clip for training are randomly sampled from the original video to ensure the robustness of the model. During inference, we select $N_r=3$ reference images for the generation process of each fashion video. To ensure that the selected images cover as many necessary details from different perspectives as possible, we utilize the human pose of reference images to guide the selection process. Specifically, we calculate the relative positions of left body keypoints and right body keypoints and divide the video frames into 3 orientation groups accordingly: front, back, and side. Finally, we randomly choose an image from each group to form the reference images.

\sssection{Detailed Architecture.} The number of down-blocks, mid-blocks, and up-blocks in the U-Net~\cite{unet} structure is 4, 1, and 4, respectively. The extra spatiotemporal attention process is included from the last down-block to the first up-block. The number of reference images $N_r$ is set to 3 to incorporate different reference perspectives. The reference and driving pose sequences are extracted by DWPose~\cite{dwpose} and rendered by OpenPose~\cite{openpose}.

\sssection{Training.} We utilize the VAE and spatial parameters from Stable Diffusion V1.5~\cite{ldm} to initialize the model. We use AdamW~\cite{adamw} to optimize the model with a learning rate of $5\times{10}^{-5}$. The videos are resized and center-cropped to $1024\times 576$. In the first stage, we train the model with a batch size of 128 for 30,000 steps. In the second stage, a 16-frame clip is sampled from the full video for training. The model is trained with a batch size of 16 for 20,000 steps.

\sssection{Inference.} We use a DDIM~\cite{ddim} sampler for 35 steps with classifier-free guidance~\cite{cfg} scale 3.5. We use a similar temporal aggregation method to~\cite{animateanyone} for long video synthesis.

\sssection{Reproducibility.} 16 NVIDIA A100 80GB GPUs are used for training. Evaluation is done under the same condition.

\begin{figure*}
  \centering
  \begin{subfigure}{\linewidth}
    \centering
    \includegraphics[width=\linewidth]{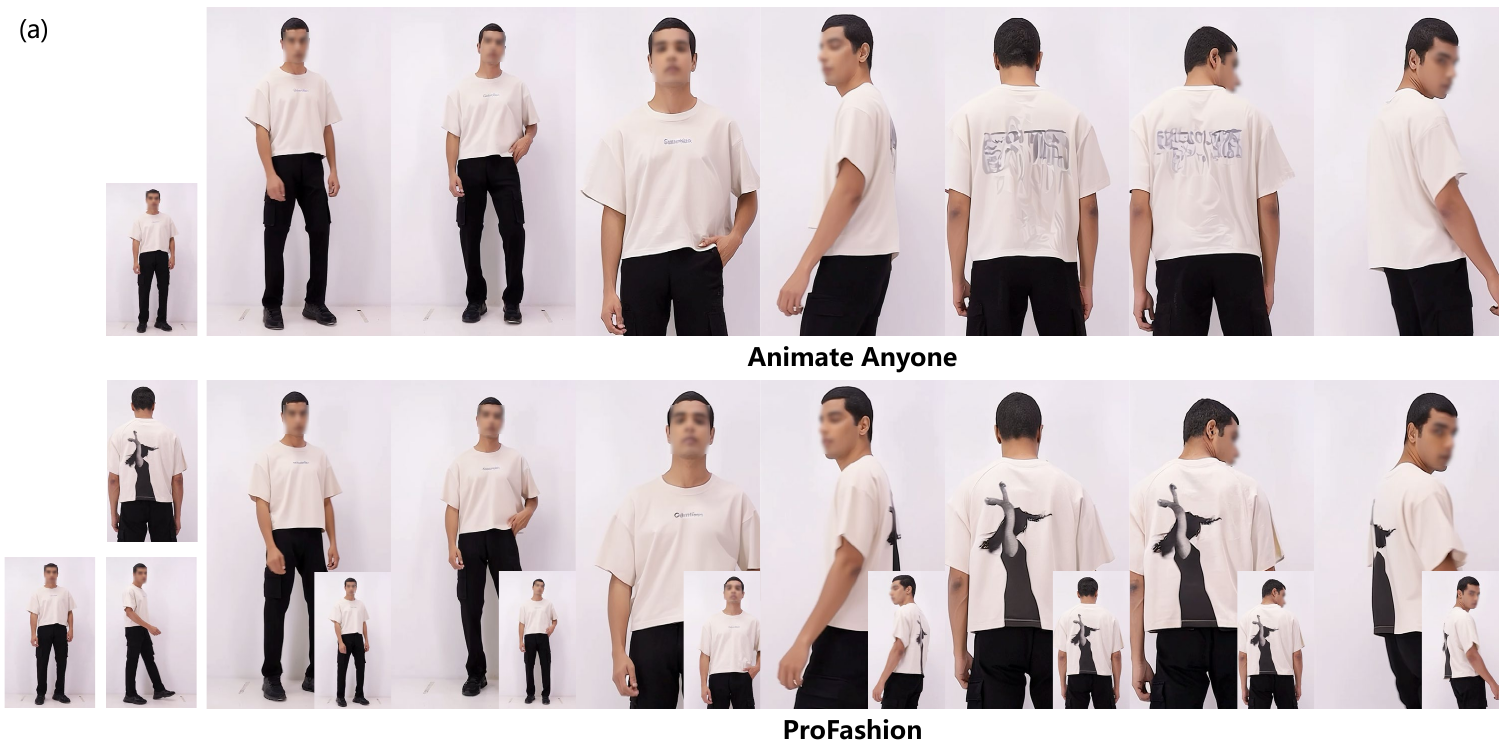}
    \label{fig:vis_mrfashion_a}
  \end{subfigure}
  \vfill
  \begin{subfigure}{\linewidth}
    \centering
    \includegraphics[width=\linewidth]{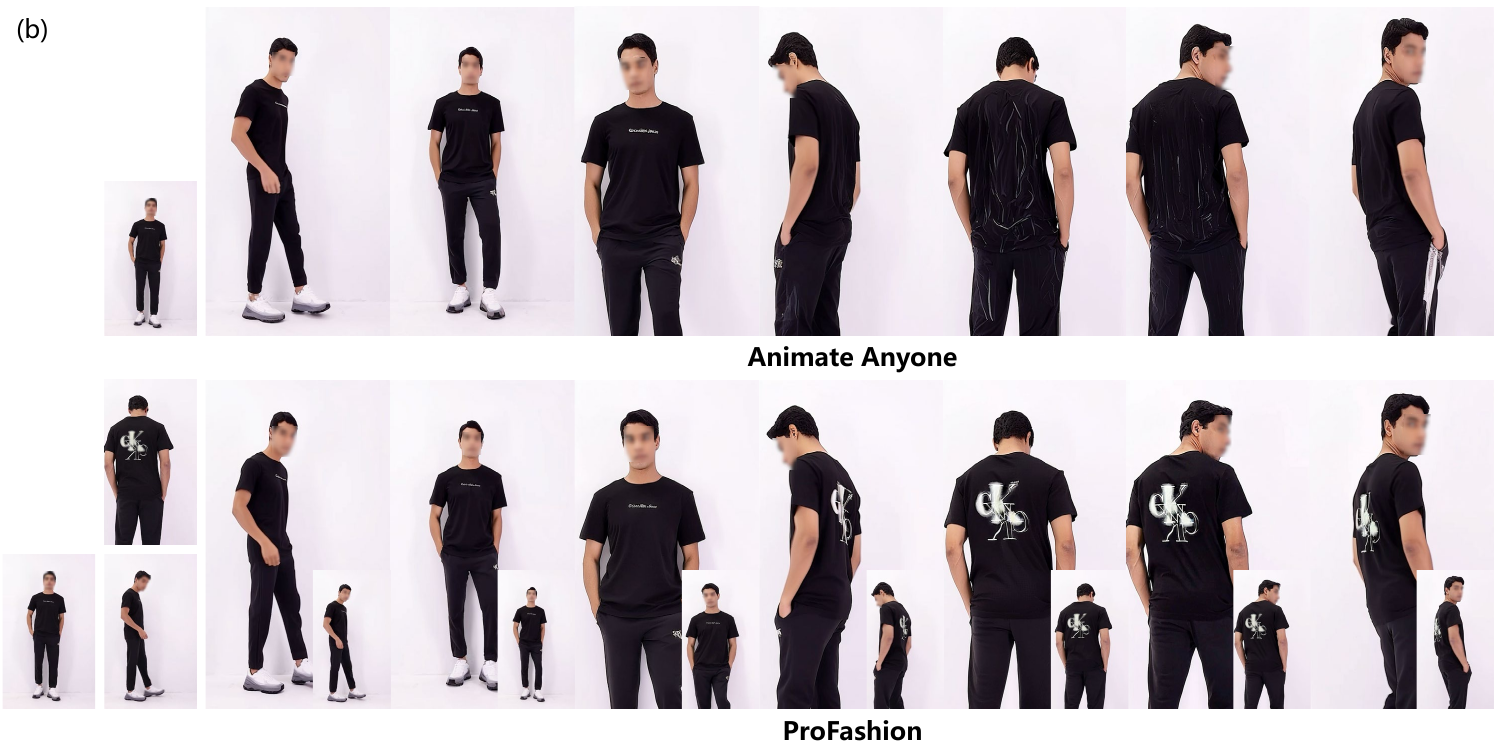}
    \label{fig:vis_mrfashion_b}
  \end{subfigure}
  \caption{Visualizations on the test split of MRFashion-7K (\sref{sec:mrfashion}). The reference images are on the left of each row. The ground truth is in the bottom-right corner of ProFashion's results.}
  \label{fig:vis_mrfashion}
\end{figure*}

\subsection{Comparisons on MRFashion-7K}
\label{sec:mrfashion}

To validate the effectiveness of our design, we compare the full model with two single-reference methods~\cite{animateanyone, champ} and four ablative designs on MRFashion-7K. The ablative designs include a single-reference baseline model (\#3), the baseline model with average pooling fusion for multiple references (\#4) and concatenation of multiple references (\#5), and the baseline model with PPA only (\#6).

\sssection{Quantitative Results.} The quantitative results are summarized in \cref{tab:mrfashion}. It can be observed that introducing multiple reference images significantly enhances the quality of generated videos compared to single-reference baselines (\#1 to \#3).
Averaging multiple references (\#4) suffers from the feature blending problem, which limits the generation quality. Although concatenating references (\#5) can better preserve garment details compared to averaging, it introduces a significant computational burden that reduces the length of training clips to 12 frames, sacrificing motion fluency. By incorporating PPA (\#6), the model achieves a significant performance boost without introducing extensive computation compared to averaging (\#3), especially on SSIM (0.838 to 0.867) and LPIPS (0.125 to 0.094). The motion smoothness of generated videos further improves by incorporating FTA (\#7 to \#6), which can be validated by the vast reduction in FVD (196.95 to 126.92, a 35.56\% improvement).

\sssection{Human Evaluation.} To ensure that the generated videos align well with the aesthetic criteria of humans, we conducted a user study by asking 13 volunteers to rate the generated fashion videos in 3 aspects: character authenticity, clothing detail, and motion fluency, with an integer score from 0 to 5. The overall quality is the average of the 3 scores mentioned before. We present the results in \cref{tab:mrfashion}. Compared to single-reference baselines, average-pooling fusion (\#4) and concatenation (\#5) do produce better results, but there is still a significant gap in meeting the user's intention. PPA (\#6) brings an observable performance boost, especially in clothing detail (3.13 to 3.69 compared to \#4). By introducing FTA (\#7), the generation quality is further enhanced, especially in motion fluency (3.12 to 3.87).

\sssection{Qualitative Results.} To demonstrate the superiority of ProFashion, we visualize synthesized videos from Animate Anyone~\cite{animateanyone} and ProFashion in \cref{fig:vis_mrfashion}. We can observe that Animate Anyone struggles with severe hallucination when generating the back side of the character, where the required information cannot be covered by a single reference image. In contrast, ProFashion is capable of generating view-consistent fashion videos under the condition of multiple reference images from different perspectives. We provide additional visual comparisons in the supplementary.

\subsection{Comparisons on UBC Fashion}
\label{sec:comp}

We conducted experiments on the UBC Fashion~\cite{dwnet_ubc} dataset for better comparison to previous state-of-the-art methods.

\begin{table}
  \centering
  \caption{Quantitative results on UBC Fashion~\cite{dwnet_ubc} dataset (\sref{sec:comp}). Results in \textbf{bold} are the best. * With sample fine-tuning. \dag~Open-source implementation.}
  \label{tab:ubc}
  \begin{tabular}{lccc}
    \toprule
    Methods & \texttt{SSIM}$\uparrow$ & \texttt{LPIPS}$\downarrow$ & \texttt{FVD}$\downarrow$ \\
    \midrule
    MRAA~\cite{mraa} & 0.749 & 0.212 & \cellcolor{Gray}253.7 \\
    TPSMM~\cite{tpsmm} & 0.746 & 0.213 & \cellcolor{Gray}247.6 \\
    PIDM~\cite{pidm} & 0.713 & 0.288 & \cellcolor{Gray}1197.4 \\
    DreamPose~\cite{dreampose} & 0.879 & 0.111 & \cellcolor{Gray}279.6 \\
    DreamPose\textsuperscript{*}~\cite{dreampose} & 0.885 & 0.068 & \cellcolor{Gray}238.7 \\
    Animate Anyone\textsuperscript{\dag}~\cite{animateanyone} & 0.871 & 0.080 & \cellcolor{Gray}125.7 \\
    ProFashion (Ours) & \textbf{0.909} & \textbf{0.068} & \cellcolor{Gray}\textbf{86.2} \\
    \bottomrule
  \end{tabular}
\end{table}

\sssection{Quantitative Results.} The quantitative comparison with state-of-the-art methods is illustrated in \cref{tab:ubc}. It can be observed that ProFashion consistently outperforms all previous methods in all metrics especially FVD, which accounts for both image and video level quality. In this metric, ProFashion surpasses the previous state-of-the-art by 39.5, which is a 31.4\% improvement.

\begin{figure}
  \centering
  \includegraphics[width=\linewidth]{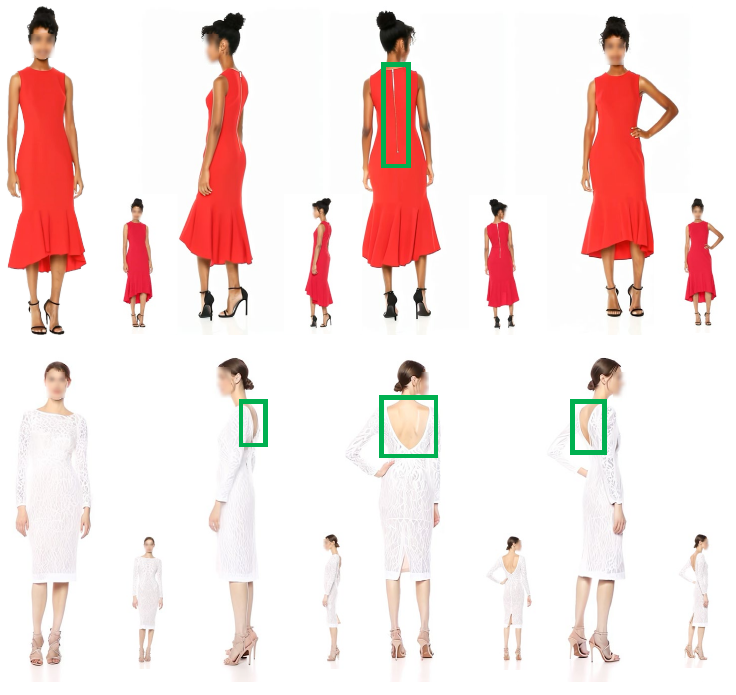}
  \caption{Visualizations on the test split of UBC Fashion~\cite{dwnet_ubc} dataset (\sref{sec:comp}). Results are generated by ProFashion with ground truth in the bottom-right corner.}
  \label{fig:vis_ubc}
\end{figure}

\sssection{Qualitative Results.} We present fashion videos generated by ProFashion on the UBC Fashion~\cite{dwnet_ubc} dataset in \cref{fig:vis_ubc}. As we can observe, ProFashion is capable of synthesizing view-consistent videos that preserve the intricate details of garments from different perspectives.

\section{Conclusion and Discussion}
\label{sec:conclusion}

In this work, we propose ProFashion, a prototype-guided fashion video generation method that effectively leverages multiple reference images as conditions to synthesize view-consistent videos, overcoming the inherent limitation of a single reference image. It introduces a fashion video generation framework with a Reference Encoder, PPA, and FPI to effectively incorporate multiple references. PPA is designed to integrate multiple reference features without significant extra computational cost. FPI is devised to further enhance motion smoothness by exploiting human keypoint motion flow. The effectiveness of ProFashion has been demonstrated by extensive quantitative and qualitative results on multiple datasets. We believe that ProFashion will promote the online retailing of clothes by providing accurate and detailed fashion videos from images at a low cost. 

\sssection{Limitations.} Despite satisfactory performance in preserving pattern-related details, ProFashion still struggles to maintain textual details on clothes. The generated videos contain distortions and blurs in textual areas. More discussions are included in the supplementary.



{
    \small
    \bibliographystyle{ieeenat_fullname}
    \bibliography{main}
}

\clearpage
\maketitlesupplementary
\appendix

This document provides extra experimental results, and corresponding analyses of ProFashion. The document is organized as follows:
\begin{itemize}
    \item \sref{sec:extra_abl} presents extra ablation results on different classifier-free guidance (CFG)~\cite{cfg} scales and another design choice of PPA.
    \item \sref{sec:generalization} demonstrates the generalization capability of ProFashion to various driving pose sequences.
    \item \sref{sec:additional_qualitative} shows additional qualitative results of ProFashion and analyzes its failure cases.
\end{itemize}

\section{Extra Ablations}
\label{sec:extra_abl}

\sssection{CFG Scale.} To explore how the CFG~\cite{cfg} scale affects the generation results, we conduct ablations on several scale factors on MRFashion-7K. The results are displayed in \cref{tab:abl_cfg}. It can be observed that the CFG scale has a prominent impact on the generation quality and needs to be appropriately tuned.

\begin{table}[htb]
  \centering
  \caption{Ablations of CFG scales (\sref{sec:extra_abl}) on MRFashion-7K. Results in \textbf{bold} are the best.}
  \label{tab:abl_cfg}
  \begin{tabular}{c|cccc}
    \toprule
    CFG Scales & \texttt{SSIM}$\uparrow$ & \texttt{PSNR}$\uparrow$ & \texttt{LPIPS}$\downarrow$ & \texttt{FVD}$\downarrow$ \\
    \midrule
    2.5 & 0.859 & 22.67 & 0.103 & \cellcolor{Gray}147.63 \\
    3.5 & \textbf{0.885} & \textbf{23.57} & \textbf{0.086} & \cellcolor{Gray}\textbf{126.92} \\
    5.0 & 0.871 & 22.23 & 0.109 & \cellcolor{Gray}196.10 \\
    7.5 & 0.859 & 22.20 & 0.105 & \cellcolor{Gray}210.20 \\
    \bottomrule
  \end{tabular}
\end{table}

\begin{figure}[htb]
    \centering
    \includegraphics[width=\linewidth]{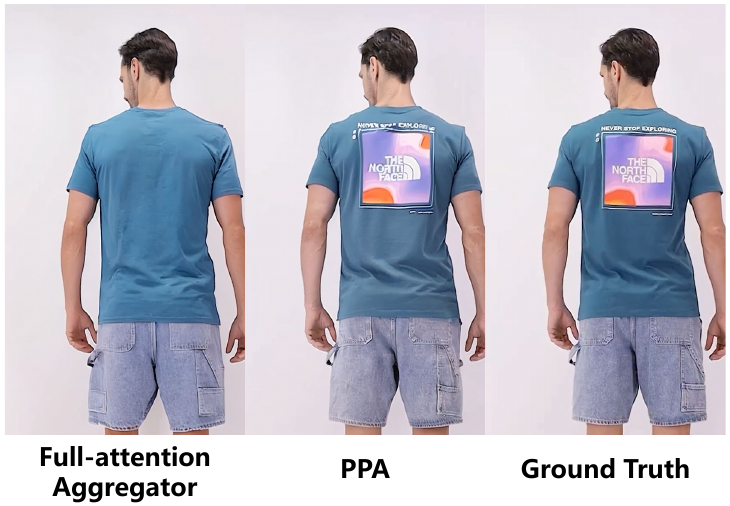}
    \caption{Comparisons of different design choices of PPA (\sref{sec:extra_abl}) on MRFashion-7K.}
    \label{fig:abl_ppa}
\end{figure}

\sssection{Design Choice of PPA.} To validate the superiority of the design of PPA, we implement another full-attention aggregator. This alternative design does not perform the average pooling operation on the spatial dimension, resulting in a full attention map between $\bm{q}_p^i$ and $\bm{k}_p^i$, which is then multiplied with the fine-grained reference features to obtain fine-grained prototypes. Such a design significantly increases the GPU memory usage, reducing the length of training clips to 12 frames. Despite faster convergence, this design fails to learn the reference selection criteria and cannot provide appropriate guidance for the generation process, leading to unsatisfactory results on MRFashion-7K (\cref{fig:abl_ppa}).

\section{Generalization Analysis}
\label{sec:generalization}

\begin{figure*}
    \centering
    \includegraphics[width=0.92\linewidth]{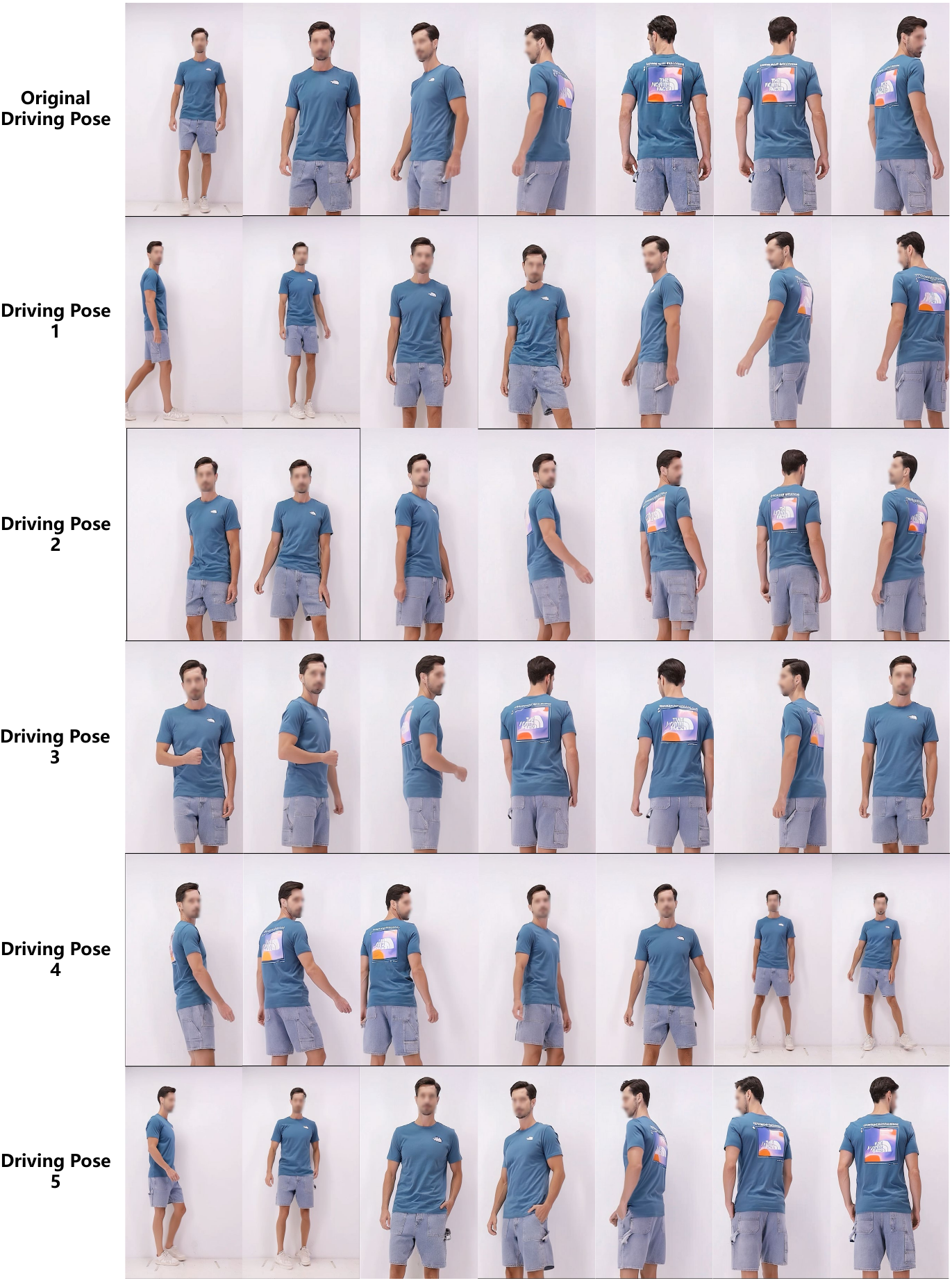}
    \caption{Fashion video generation results on MRFashion-7K with different driving pose sequences (\sref{sec:generalization}).}
    \label{fig:cross_ref}
\end{figure*}

To better demonstrate the generalization capability of ProFashion, we conduct fashion video synthesis on MRFashio-7K conditioned by driving pose sequences from other videos than the reference. Results are shown in \cref{fig:cross_ref}. As observed, ProFashion achieves satisfactory results, maintaining view consistency and motion smoothness.

\section{Additional Qualitative Results}
\label{sec:additional_qualitative}

\begin{figure*}
  \centering
  \begin{subfigure}{\linewidth}
    \centering
    \includegraphics[width=\linewidth]{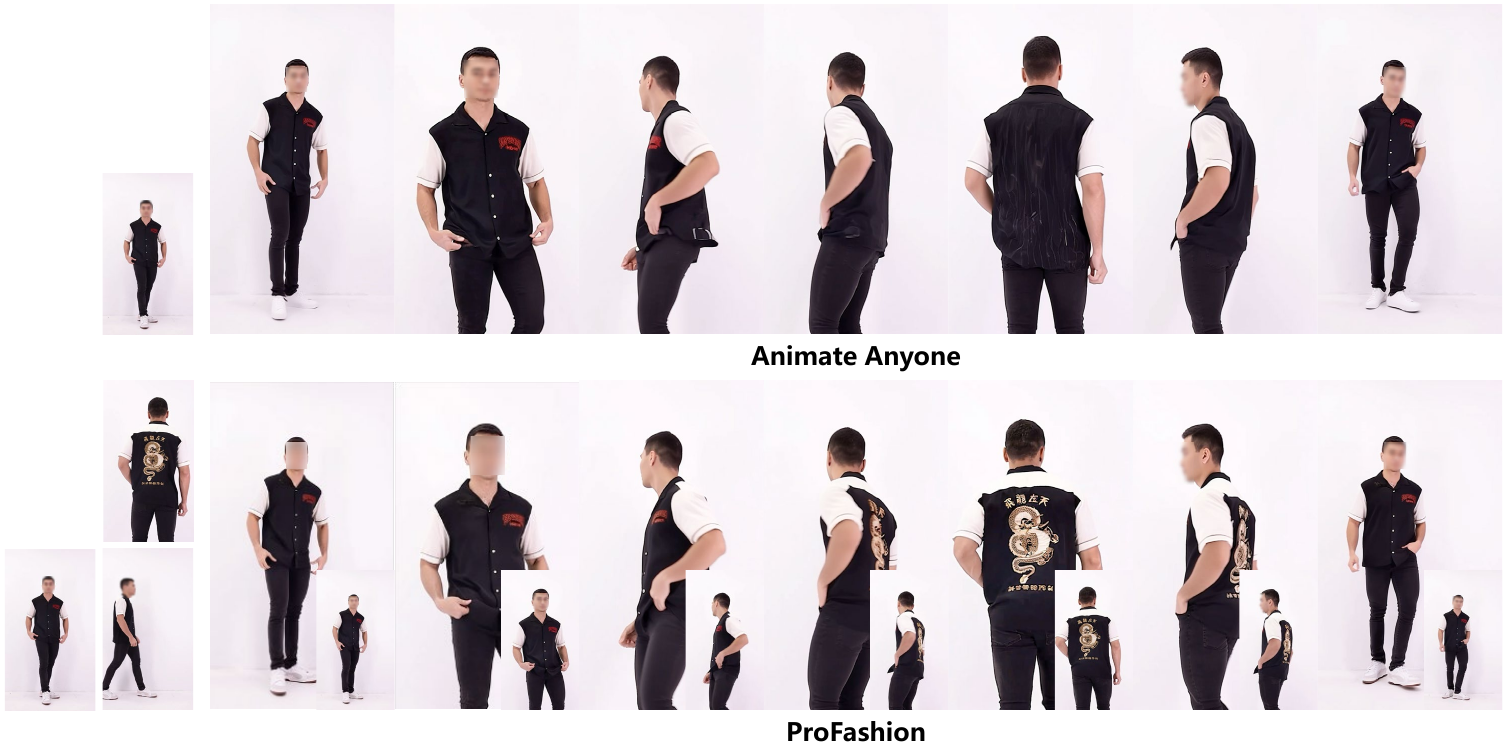}
  \end{subfigure}
  \vfill
  \begin{subfigure}{\linewidth}
    \centering
    \includegraphics[width=\linewidth]{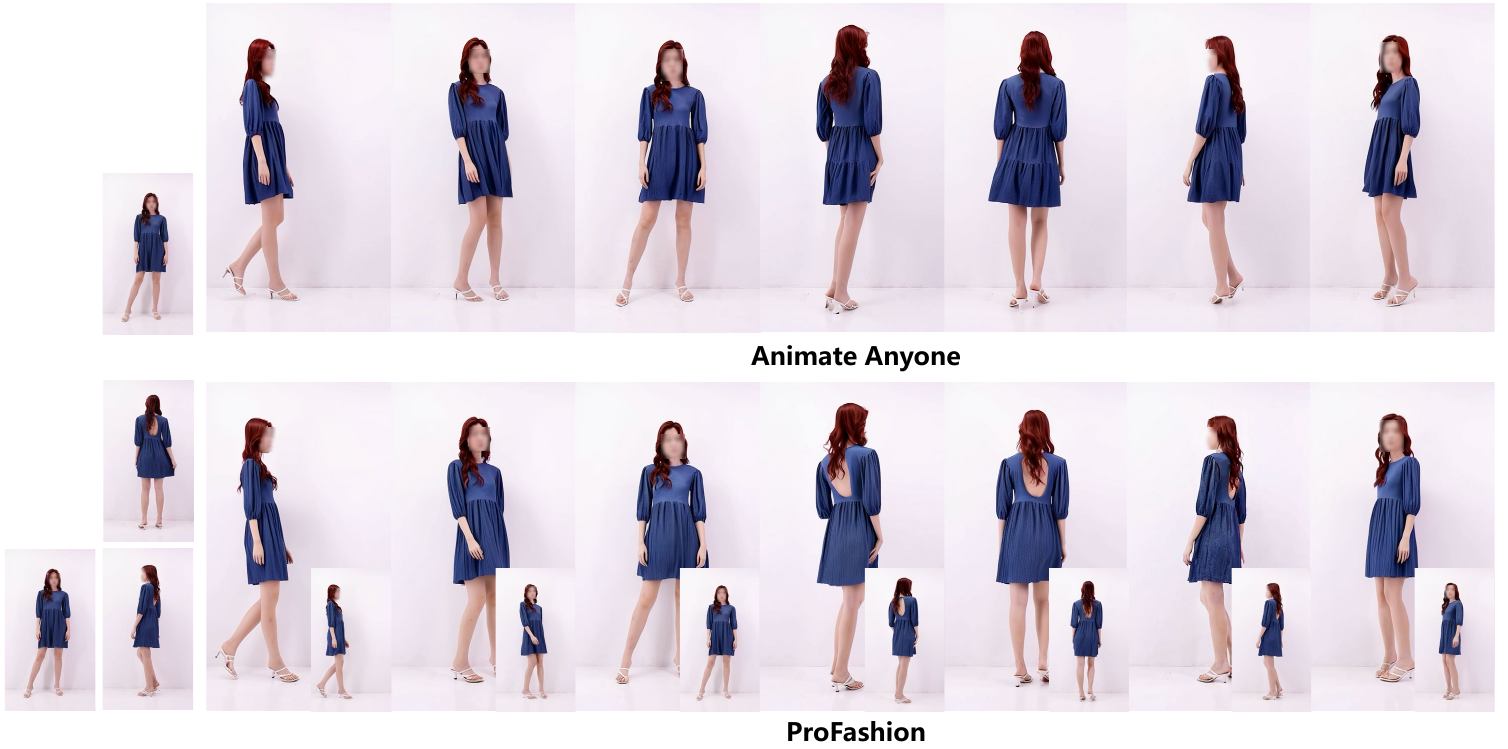}
  \end{subfigure}
  \caption{More visual comparisons with Animate Anyone~\cite{animateanyone} on the test split of MRFashion-7K (\sref{sec:additional_qualitative}). The reference images are on the left of each row. The ground truth is in the bottom-right corner of ProFashion's results.}
  \label{fig:supp_vis_mrfashion}
\end{figure*}

\begin{figure}
  \centering
  \includegraphics[width=\linewidth]{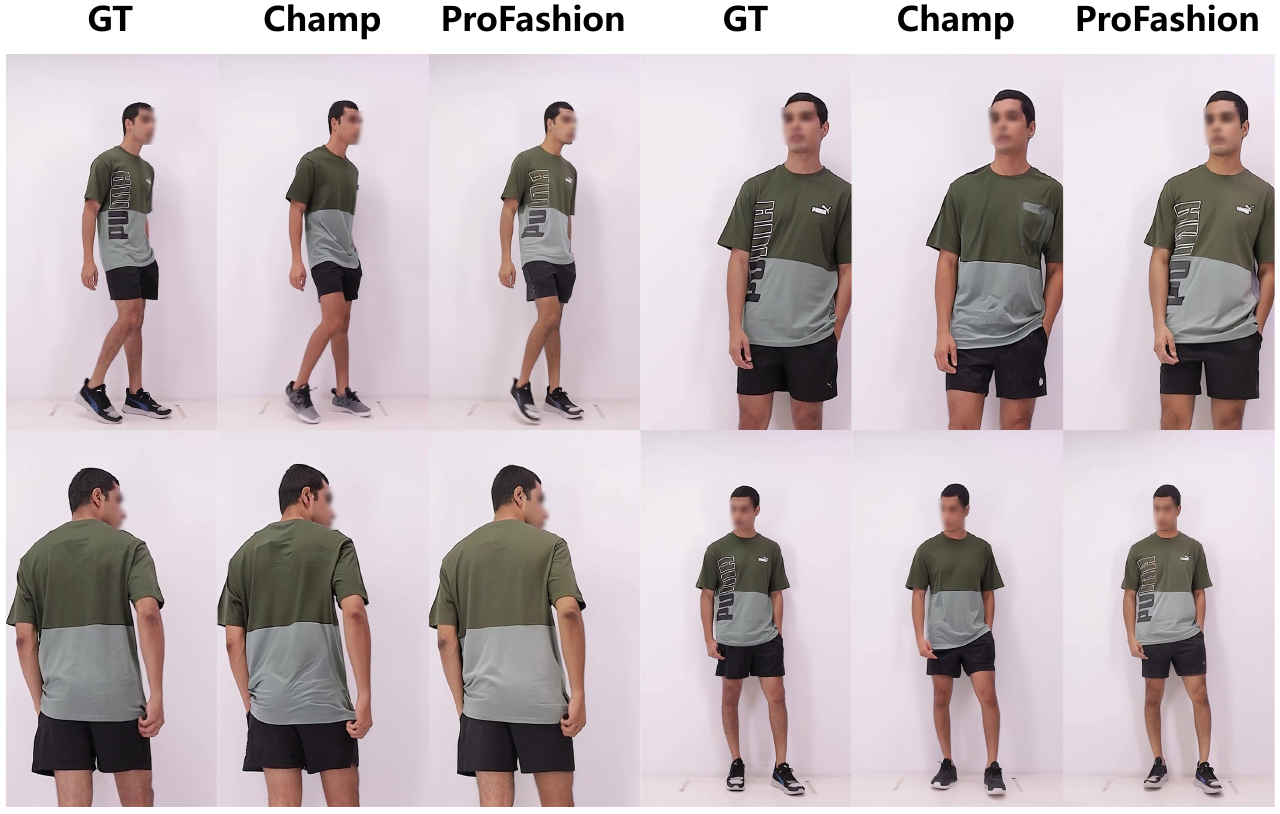}
  \caption{Visual comparisons with Champ~\cite{champ} on the test split of MRFashion-7K (\sref{sec:additional_qualitative}).}
  \label{fig:supp_vis_champ}
\end{figure}

We provide more qualitative results on MRFashion-7K in \cref{fig:supp_vis_mrfashion,fig:supp_vis_champ} and the supplementary video to illustrate the effectiveness of ProFashion. Compared to single-reference baselines~\cite{animateanyone, champ}, ProFashion is capable of genuinely reproducing garment details from multiple reference images into a smooth fashion video containing various perspectives of the character.

\begin{figure}
    \centering
    \includegraphics[width=\linewidth]{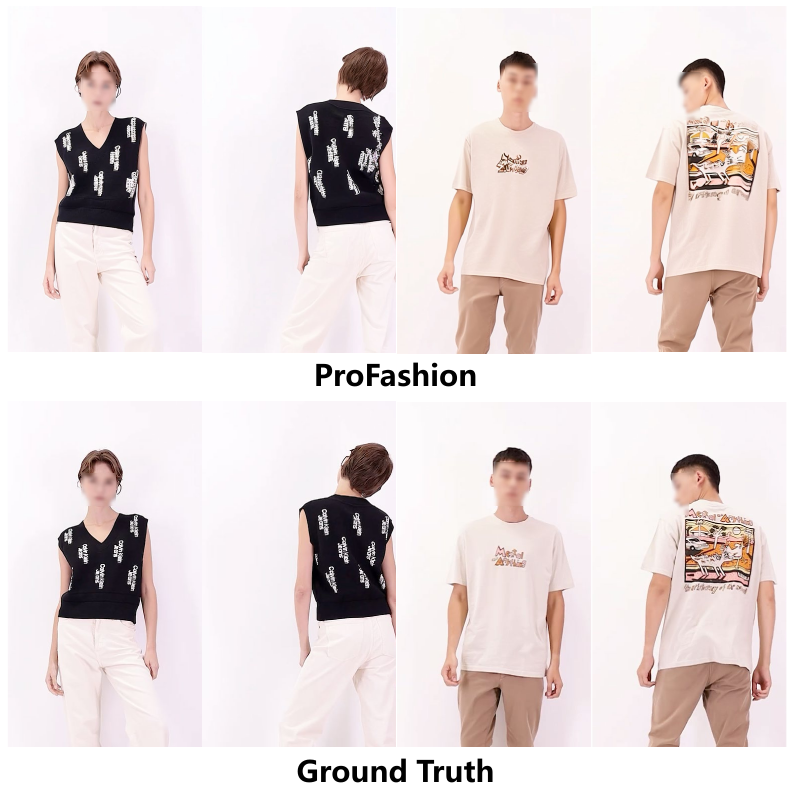}
    \caption{Failure cases concerning textual details (\sref{sec:additional_qualitative}) on MRFashion-7K.}
    \label{fig:failure}
\end{figure}

\sssection{Failure Cases.} Despite its effectiveness, ProFashion falls short in synthesizing texts on clothes. As \cref{fig:failure} illustrates, ProFashion struggles to generate clear and recognizable letters in these cases. In contrast, significant distortions and blurs are introduced in textual areas, limiting the application of ProFashion to garments with extensive textual details. The inability to neatly handle textual details can be explained by the blending of reference features in Eq.~(8), which can potentially be addressed by preserving the original features of textual areas in our future work.


\end{document}